\documentclass[runningheads]{llncs}
\usepackage[T1]{fontenc}
\usepackage{graphicx}
\usepackage[english]{babel}

\usepackage{amsmath}
\usepackage{graphicx}
\usepackage{url}
\title{UruBots \\ RoboCup@Work \\ Team Description Paper}
\author{Hiago Sodre, Juan Deniz, Pablo Moraes, William Moraes, Igor Nunes, \\ Vincent Sandin, Ahilen Mazondo, Santiago Fernandez, Gabriel da Silva, \\ Mónica Rodríguez, Sebastian Barcelona, Ricardo Grando}

\authorrunning{H. Sodre et al.}
\institute{
Robotics and Artificial Intelligence Lab \\
Technological University of Uruguay \\
Rivera, Uruguay \\
Email: \url{urubots.itrn@utec.edu.uy} \\
Website: \url{urubots.uy}\\
}

\begin{document}
\maketitle 

% \begin{centering}
% Robotics and Artificial Intelligence Lab \\
% Technological University of Uruguay \\
% Rivera, Uruguay \\
% Email: \url{urubots.itrn@utec.edu.uy} \\
% Website: \url{urubots.uy}\\
% \end{centering}

\begin{abstract}
This work presents a team description paper for the RoboCup @work League. Our team, UruBots, has been developing robots and projects for research and competitions in the last three years, attending robotics competitions in Uruguay and around the world. In this instance, we aim to participate and contribute to the RoboCup @Work category, hopefully making our debut in this prestigious competition. For that, we present an approach based on the Limo robot, whose main characteristic is its hybrid locomotion system with wheels and tracks, with some extras added by the team to complement the robot’s functionalities. Overall, our approach allows the robot to efficiently and autonomously navigate a @work scenario, with the ability to manipulate objects, perform autonomous navigation, and engage in a simulated industrial environment.
\end{abstract}

\section{Introduction}

UruBots is a robotics competition team created in 2022 at the Technological University of Uruguay. The team has been developing projects and attending competitions in the areas of mobile robotics, artificial intelligence, computer vision, and other currently relevant topics. Within the field of competitions, we have already competed in categories such as Mission Impossible, Autonomous Cars, HuroCup, and Emergency Indoor Service categories of FIRA RoboWorldCup \cite{moraes2024urubots} y \cite{sodre2024urubots}. Part of the team has already performed research with mobile robotics for industrial-related tasks \cite{deniz2024real}, \cite{grando2021deep}, \cite{grando2022double}, \cite{grando2022mapless} y \cite{grando2024improving}.

\begin{figure}[!t]
\centering
\includegraphics[width=\linewidth]{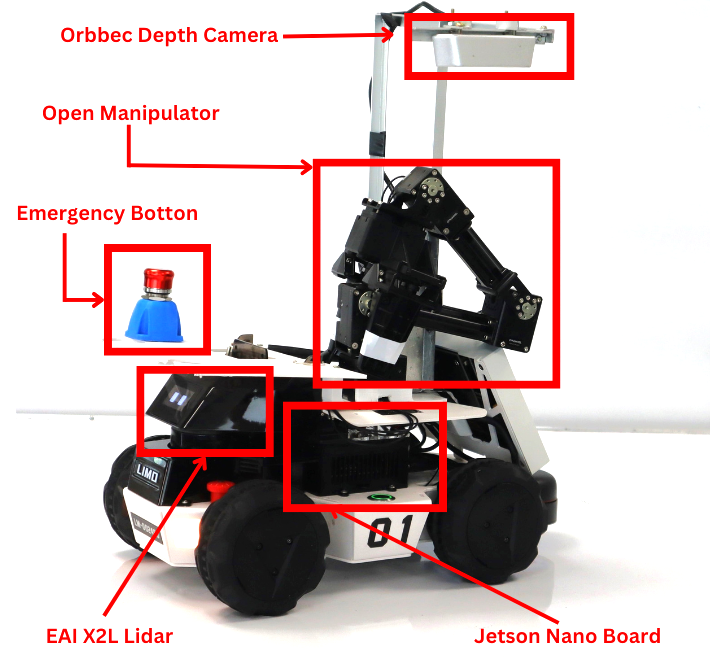}
\caption{Proposed Robotic Platform.}
\label{fig:robotPhoto}
\end{figure}

Our proposed robotic system for the Robocup@work league, as shown in Figure \ref{fig:robotPhoto}, has a differential mobile base equipped with an open manipulator, a depth camera, a 2D Lidar and a Jetson Nano processing board. The robot is capable of navigating autonomously, detecting objects, and manipulating them in the workplace environment.

% Our goal is to optimize autonomous navigation and decision-making in complex environments. We will implement advanced perception and control algorithms to detect objects and adverse situations and manipulate different materials. Using interfaces such as ROS with the Limo robot, it is possible to map such environments and provide precise navigation over them.

\subsection{Scientific Publications}

So far, part of our team has already engaged in research related to situation awareness for mobile robotics in industrial-related environments \cite{deniz2024real}, \cite{sodre2024urubots}.  
At Juan et al. \cite{deniz2024real}, we present a Real-time Robotic Situational Awareness for Accident Prevention in Industry, where we developed a methodology based on the Locobot robot with the YOLO framework to detect dangerous situations in a simulated industrial environment and provide real-time information for the workplace. With this approach, it is possible to interact with an end user - in this case, a worker, for example - and help avoid accidents. We compared different YOLO versions (YOLOv8 and YOLOv5) with autonomous navigation so the robot could work autonomously. Overall, this study improves knowledge in navigation and object detection, contributing to the development of safer human-robot interaction (HRI) systems, which will be explored in this article in the context of RoboCup@Work.

Some team members have also developed articles related to object manipulation and navigation with mobile robots, utilizing artificial intelligence techniques such as YOLO and others \cite{montenegro2018neural}, \cite{moraes2024behavior}. Additionally, research has been conducted in Deep Reinforcement Learning for Mapless Navigation, which provides significant advancements in autonomous movement without the need for predefined maps \cite{grando2021deep}, \cite{grando2022double}, \cite{grando2024improving}. These studies contribute to improving robotic perception, decision-making, and adaptability in dynamic environments in the field of mobile robotics.

\section{Mechanical and Hardware Description}

This section presents the characteristics of the base robot used, and the improvements made to the robot to meet the competition demands. Our system is based on a mobile robotics platform embedded with a processing unit and a vision platform, plus an attached manipulator. We also added an emergency stop button above the platform, following the rules of the competition.

\subsection{Robotic Platform}

Our mechanical and hardware system is based on the Limo robot, which was created by AgileX. It integrates the NVIDIA Jetson Nano alongside the EAI X2L LiDAR and ORBBEC Dabai depth camera. This enables robust environmental perception and intelligence for autonomous navigation, obstacle avoidance, and visual recognition applications.

The robotic platform incorporates a ranging sensor, the EAI X2L LiDAR, ideal for autonomous mapping and localization, together with an ORBBEC Dabai depth camera, which facilitates navigation planning and dynamic obstacle avoidance, allowing for simultaneous localization and mapping. Its design allows it to work in multiple driving modes, including omnidirectional, tracked, Ackerman, and four-wheel differential options, allowing it to adapt to a wide range of environments, both indoors and outdoors. 

The robot is powered by a 10,000 mAh battery that provides up to 2.5 hours of continuous autonomy, improving its operability and stability in prolonged applications. It is also compatible with ROS 1, offering support for programming languages such as Python and C++, encouraging the development of new robotic applications.

In terms of physical dimensions, the LIMO measures 322 mm long, 220 mm wide, and 251 mm high, weighing approximately 4.8 kg. In Table \ref{tab:SystemRobot1} below, it is possible to view more technical information about the robot.

\begin{table}[h]
\renewcommand{\arraystretch}{1}
 \tabcolsep=0.1cm
\caption{Technical specifications of our robotic platform}
\label{tab:SystemRobot1}
\centering
\begin{tabular}{|l|r|}
\hline
Attribute & Value \\ \hline
Name & Limo  \\
Overall dimension  & 322*220*251mm \\
Wheelbase & 200mm \\
Tread  & 175mm \\
Dead load  & 4.8kg \\
Payload & 4kg \\
Minimum ground clearance & 24mm \\
Drive type & Hub motor(4x14.4W) \\
No-load max. speed  & 1m/s \\
LiDAR & EAI X2L\\
Depth Camera & ORBBEC Dabai\\
IPC & Jeston Nano\\
Battery  & 5.2AH 12V\\
Operating System  & Ubuntu 18.04\\
ROS Version & ROS1 Melodic\\
App Control Range & 10m\\
Control Method & Mobile App/Command Control\\

\hline
\end{tabular}
\end{table}

% \newpage
\subsection{Processing and Communication platform}

NVIDIA Jetson Nano, used for processing and controlling our robotics solutions, is a compact platform with a 128-core CUDA GPU based on the Maxwell architecture. For storage, it uses a microSD card, although the B01 version provides NVMe support through the PCIe M.2 port, which significantly increases data access speed. Its 128 CUDA core GPU and compatibility can be integrated with sensors such as LiDAR, cameras, and encoders, facilitating environmental awareness and route planning. In addition, it has low energy consumption and supports deep learning frameworks such as TensorFlow and PyTorch.

\begin{figure}[h]
\centering
\includegraphics[scale=.11]{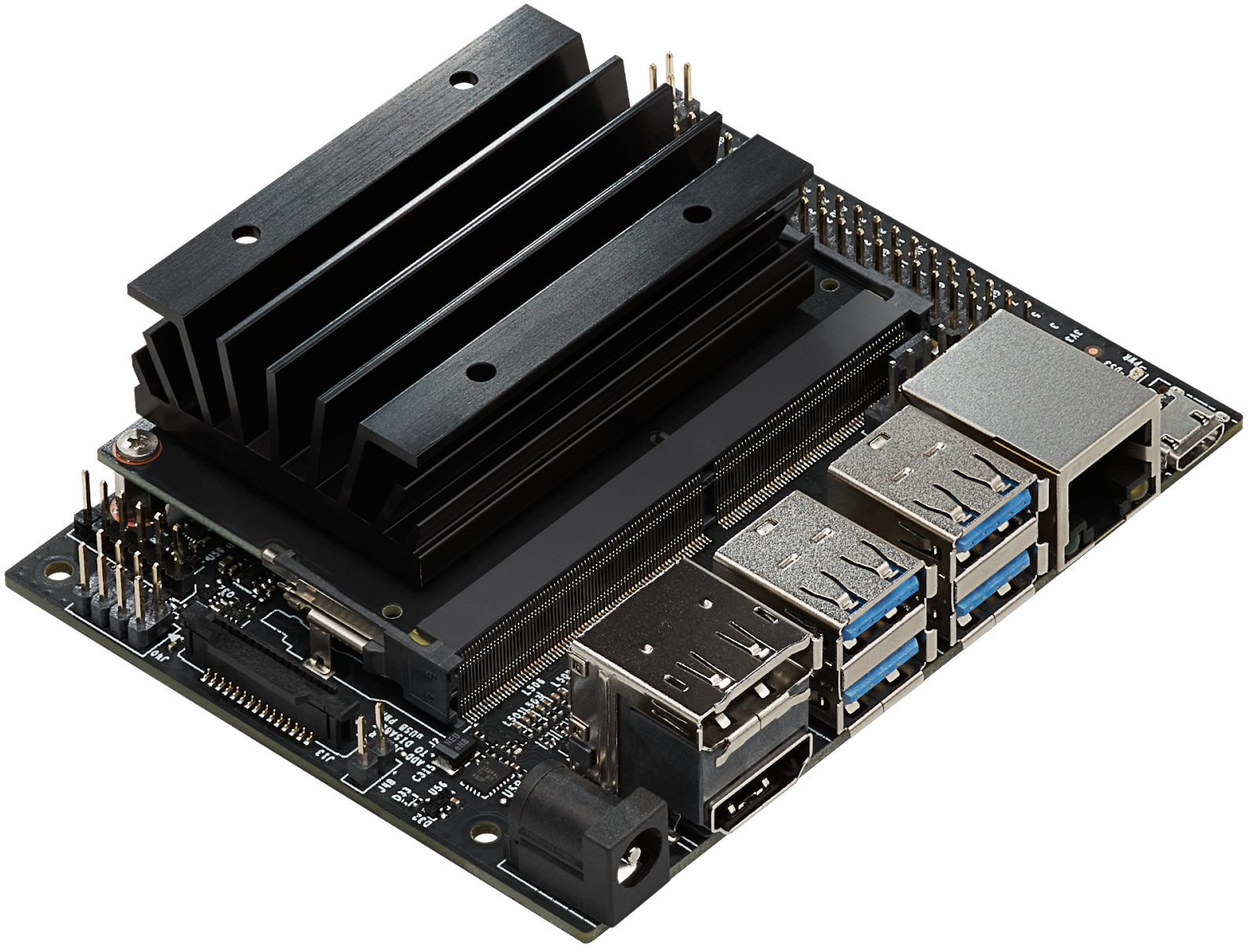}
\caption{Nvidia Jetson Nano. \cite{Nano}}
\label{fig:processing}
\end{figure}

\begin{figure}[h]
\centering
\includegraphics[scale=.55]{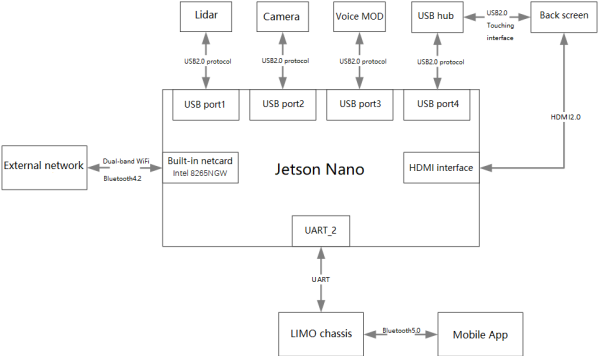}
\caption{Communication topology of the Limo Robot. }
\label{fig:communication}
\end{figure}

The LIMO chassis incorporates a Bluetooth 5.0 module that allows connection with a mobile application for remote control. It connects directly to the Nano board through a UART interface, allowing it to be controlled. In addition, it has a USB HUB that offers two USB ports and one Type-C port, all compatible with the USB 2.0 protocol. The rear screen, with a touch function, is connected to the USB HUB via the USB 2.0 interface. Figure \ref{fig:communication} shows the communication topology of the processing board.

\subsection{Manipulation Platform}

A manipulator was added to the Limo robot system to perform the requested manipulation tasks. We used a ROBOTIS OpenManipulator-X \cite{robotis-open}, a robotic arm designed for education and teaching purposes, providing mobile manipulation capabilities fully integrated with the robot operating system (ROS). Its design to implement in robots like the Turtlebot facilitates its adaptation to similar platforms based on ROS. In this case, it was used in the base above the Limo robot for manipulation tasks demanded at the competition.

\begin{figure}[h]
\centering
\includegraphics[scale=.3]{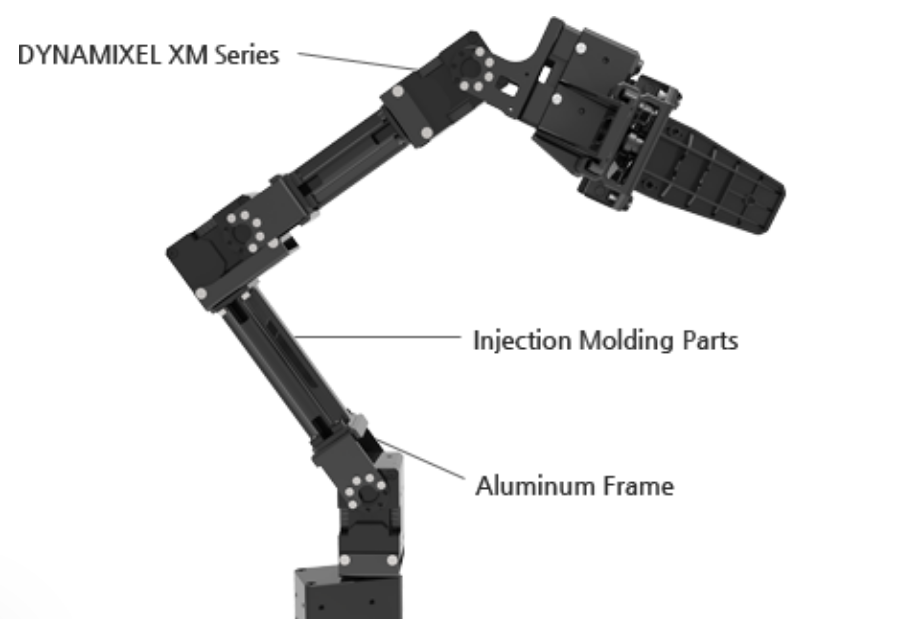}
\caption{ROBOTIS OpenManipulator-X \cite{robotis-open}. }
\label{fig:openmanipulator}
\end{figure}

With a reach of 380 mm, the OpenManipulator is suitable for robotics research and development tasks. In addition, it can operate independently using an optional base plate (Base Plate-02), which expands its versatility for different projects. 

\subsection{Vision Platform}

\begin{figure}[h]
\centering
\includegraphics[scale=.4]{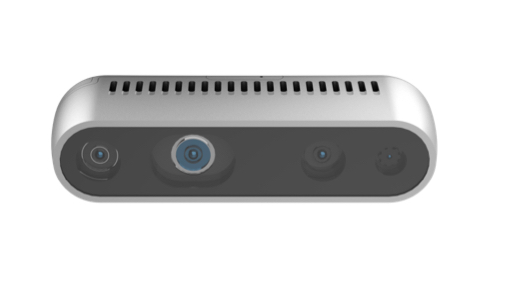}
\caption{ORBBEC Dabai Deep Camera. }
\label{fig:dabai}
\end{figure}

To facilitate object detection and manipulation for the proposed tasks, an ORBBEC Dabai depth camera was mounted on a suspended support, allowing detection and segmentation of objects to be manipulated by the OpenManipulator. The ORBBEC Dabai camera is a depth camera based on binocular structured light 3D imaging technology. The figure \ref{fig:dabai} shows the sensor.

Regarding image quality, the depth map offers resolutions of 640x400 pixels at 30FPS and 320x200 pixels at 30FPS. Resolution for color images reaches 1920x1080 pixels at 30FPS, with additional options of 1280x720 and 640x480 pixels at the same frame rate. Its accuracy is 6 mm at a distance of 1 meter, with 81 percent of the viewing area contributing to the accuracy calculations.

It has a distance of 40 mm between the optical centers of its infrared cameras. Its depth range covers from 0.3 to 3 meters, allowing precise captures in various applications. Power consumption is efficient, with an average of less than 2W, reaching a maximum of 5W at the moment of laser activation, whose duration is 3 milliseconds, while in standby mode, consumption is less than 0.7W.

\section{Software Description}

Our software system is based on the Robot Operational System (ROS). We used the built in packages of the Limo robot to perform the Simultaneos Localization and Mapping (SLAM), as well as the Open Manipulator and ORBBEC Dabai packages and our planing system.  

% \begin{figure}[h]
% \centering
% \includegraphics[scale=.55]{figs/roscomunication.PNG}
% \caption{ROS Communication Mechanism. }
% \label{fig:ros}
% \end{figure}

\subsection{Navigation}

For the navigation we use Limo's navigation system provided by its manufacturer \footnote{$https://github.com/agilexrobotics/limo_ros$}. It has its own packages to perform the movement and odometry and bring up the Lidar and the Depth camera. 

\begin{figure}[h]
\centering
\includegraphics[scale=.27]{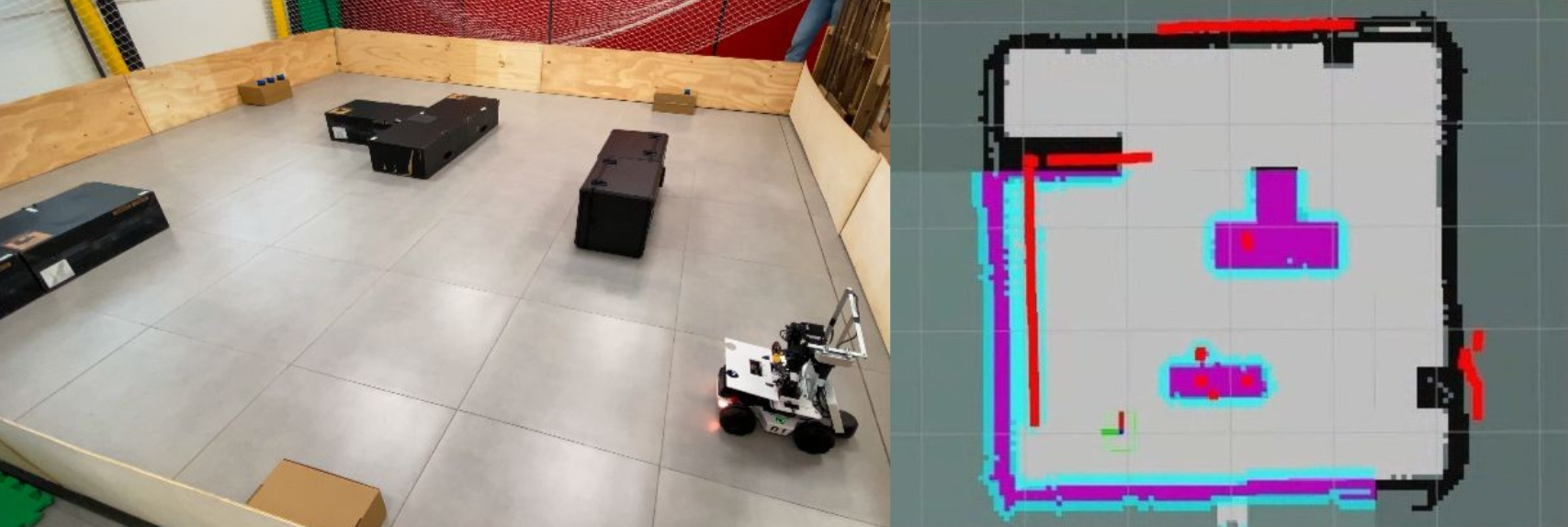}
\caption{Simulated scenario and generated map.}
\label{fig:navigation}
\end{figure}

We generate the map of a proposed scenario using its cartographer node \footnote{$https://github.com/agilexrobotics/limo_ros/blob/master/limo_bringup/launch/limo_cartographer.launch$}, which takes into consideration the information from its odometry and lidar output. Figure \ref{fig:navigation} shows our simulated @work scenario (left) with the robot located at the generated map (right).

Once the mapping was manually finished, we performed the SLAM using the robot's navigation system \footnote{$https://github.com/agilexrobotics/limo_ros/blob/master/limo_bringup/launch/limo_navigation_diff.launch$}. 

\subsection{Perception}

\begin{figure}[h]
\centering
\includegraphics[scale=.5]{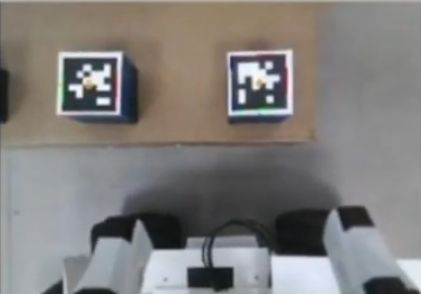}
\caption{Detection of objects based on the April Tags.}
\label{fig:apriltag}
\end{figure}

Our perception system uses the ORBBEC Dabai camera, which was installed above the robot, to assist the manipulator with the manipulation tasks. Our perception system so far is composed of the April Tag packages \footnote{$https://github.com/AprilRobotics/apriltag_ros$} to perform the detection of marks in objects in the scenario. As shown in Figure \ref{fig:apriltag}, it is used to detect the position and orientation of the objects in the scenario, feeding it to the manipulator package. For the competition, we aim also to bring a more advanced approach based on the YOLO neural network so as to detect objects that are not tagged.

\subsection{Task Planner}

\begin{figure}[h]
\centering
\includegraphics[scale=.35]{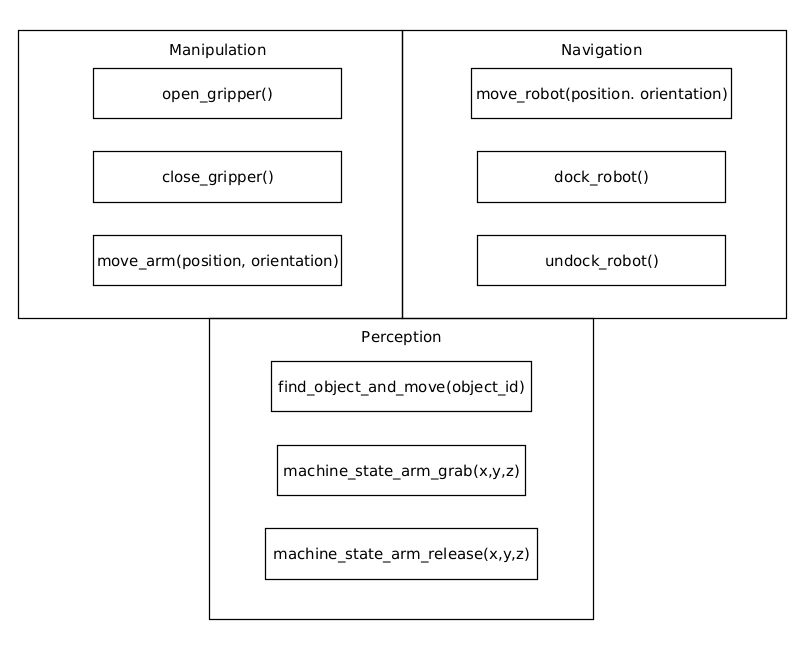}
\caption{Task Planner of our proposed system. }
\label{fig:machinestate}
\end{figure}

For the last, our planning system is a ROS package that brings up all the other packages, including a node\footnote{$https://github.com/UruBots/solverbot-work$} responsible for performing the intelligence of the proposed tasks. Figure \ref{fig:machinestate} shows the structure of our node developed so far with its main functions. 

\begin{figure}[!h]
\centering
\includegraphics[scale=.25]{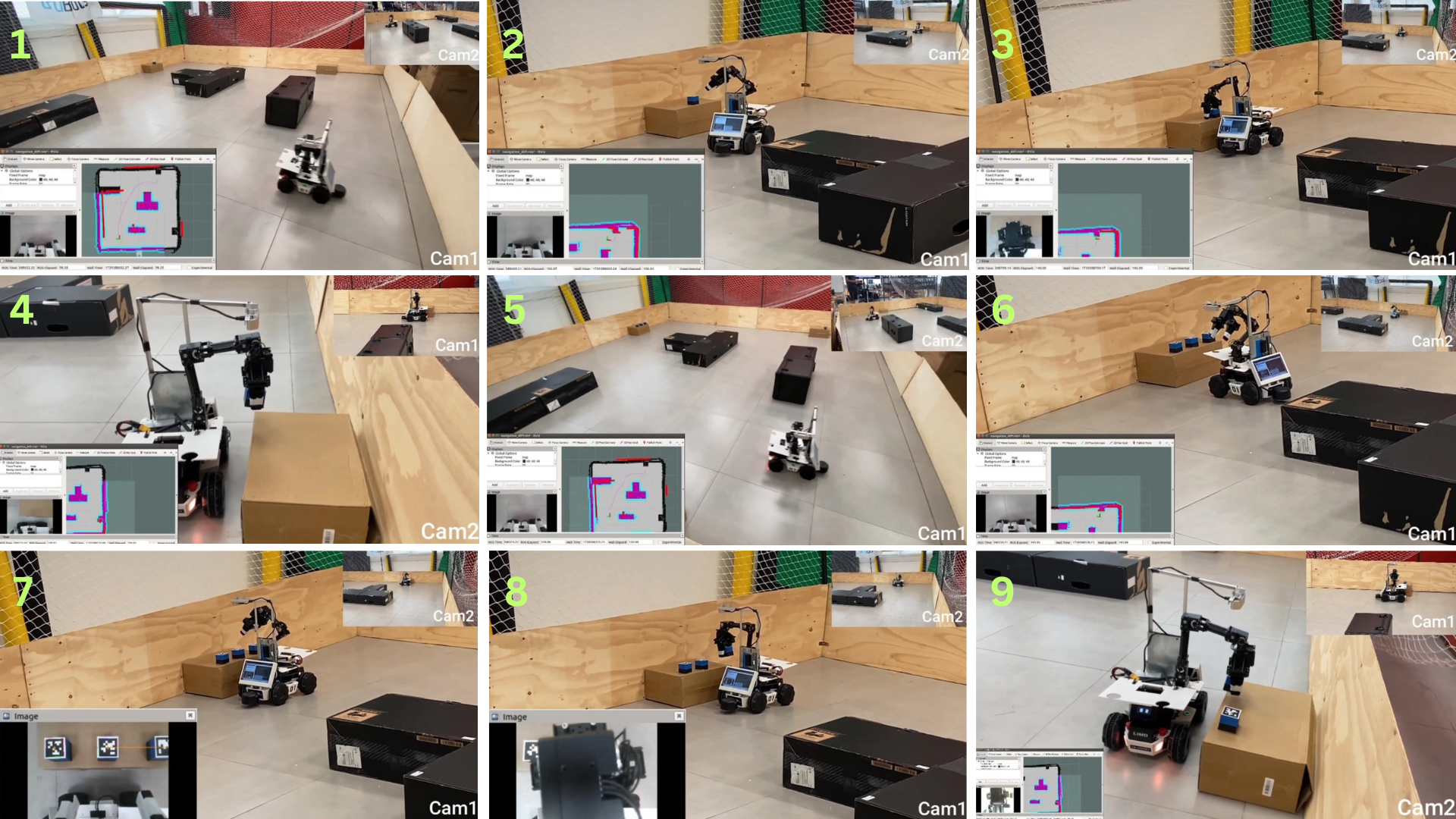}
\caption{Timeframe of our qualification experiment.}
\label{fig:timeframe}
\end{figure}

It has basic functions, such as open and close the manipulator's gripper, move the arm to a specific cartesian position and orientation and  move the robot autonomously to a specific position and orientation on the generated map. We also created a docking and undocking abilities, which, once the robot is close to base of objects, it aligns the robot correctly so the objects can be detected by the vision platform. 

For the last, we implemented a function to perform a sequence of movements for the manipulator, so it grabs a desired object and places it within the robot to be placed somewhere else.

\section{Qualification Experiment}

For this paper, we performed an experiment to show our proposed system capability to perform a @work task. We set our robot to navigate from a starting position, with a map manually generated before in a simulated scenario. Our scenario consists of a 5x5 meters field, walled on all four sides. Inside the scenario we places some obstacles and also three bases where objects with April Tags were placed. 

We set our robot to navigate to one of the bases, perform the docking, detect and grab specifically one object, and navigate and place the object in another base. We did that for a second time, where our system should also detect a specific object among others placed in the base. Figure \ref{fig:timeframe} shows a timeframe of our robotic platform performing the experiment. A qualification video\footnote{$https://www.youtube.com/watch?v=qb95qSTk60k$} was also done to show the vehicle's performance.

\section{Conclusion}

In this paper, we presented our proposed approach for the RoboCup@Work 2025 competition. Our system presents abilities to map, navigate autonomously, detect, and manipulate objects in an @work scenario. We look forward to having the opportunity to qualify for the RoboCup@Work competition, presenting our solution, and learning from the league for future participation and developments.

\section*{Acknowledgment}

We gratefully acknowledge the continued support of the team by the Technological University of Uruguay.

\bibliographystyle{abbrv}
\bibliography{sample}

\end{document}